\documentclass[a4paper]{article}

\usepackage{INTERSPEECH2021}

\usepackage{amsfonts}
\usepackage{amsmath}
\usepackage{xcolor, graphicx}
\usepackage{multirow}
\usepackage{esvect}

\makeatletter
\renewcommand{\section}{\@startsection
  {section}%
  {1}%
  {}%
  {-0.3\baselineskip}%
  {0.2\baselineskip}%
  {}}%

\renewcommand{\subsection}{\@startsection
  {subsection}%
  {2}%
  {}%
  {-0.1\baselineskip}%
  {0.05\baselineskip}%
  {}}%

\renewcommand{\subsubsection}{\@startsection
  {subsubsection}%
  {3}%
  {}%
  {-0.2\baselineskip}%
  {0.1\baselineskip}%
  {}}%

\g@addto@macro\normalsize{%
  \setlength\abovedisplayskip{5pt plus 2pt minus 2pt}
  \setlength\belowdisplayskip{5pt plus 2pt minus 2pt}
  \setlength\abovedisplayshortskip{4pt plus 2pt minus 2pt}
  \setlength\belowdisplayshortskip{4pt plus 2pt minus 2pt}
}

\makeatother

\usepackage{cleveref}
\Crefname{equation}{Eq.}{Eqs.}
\Crefname{figure}{Fig.}{Figs.}
\Crefname{tabular}{Tab.}{Tabs.}

\DeclareMathOperator*{\argmax}{arg\,max \hspace{2mm}}

\newcommand\numberthis{\addtocounter{equation}{1}\tag{\theequation}}

\renewcommand{\vec}{\vv}

\title{Acoustic Data-Driven Subword Modeling for End-to-End Speech Recognition}
\name{Wei Zhou$^{1,2}$, Mohammad Zeineldeen$^{1,2}$, Zuoyun Zheng$^{1}$, Ralf Schl\"uter$^{1,2}$, Hermann Ney$^{1,2}$}
\address{
$^1$Human Language Technology and Pattern Recognition, Computer Science Department,\\
  RWTH Aachen University, 52074 Aachen, Germany\\
$^2$AppTek GmbH, 52062 Aachen, Germany}
\email{\{zhou, zeineldeen, schlueter, ney\}@cs.rwth-aachen.de, zuo.yun.zheng@rwth-aachen.de}

\begin{document}

\maketitle
\begin{abstract}
Subword units are commonly used for end-to-end automatic speech recognition (ASR), while a fully acoustic-oriented subword modeling approach is somewhat missing.
We propose an acoustic data-driven subword modeling (ADSM) approach that adapts the advantages of several text-based and acoustic-based subword methods into one pipeline.
With a fully acoustic-oriented label design and learning process, ADSM produces acoustic-structured subword units and acoustic-matched target sequence for further ASR training.
The obtained ADSM labels are evaluated with different end-to-end ASR approaches including CTC, RNN-Transducer and attention models. Experiments on the LibriSpeech corpus show that ADSM clearly outperforms both byte pair encoding (BPE) and pronunciation-assisted subword modeling (PASM) in all cases.
Detailed analysis shows that ADSM achieves acoustically more logical word segmentation and more balanced sequence length, and thus, is suitable for both time-synchronous and label-synchronous models.
We also briefly describe how to apply acoustic-based subword regularization and unseen text segmentation using ADSM.
\end{abstract}
\noindent\textbf{Index Terms}: subword modeling, speech recognition

\section{Introduction \& Related Work}
End-to-end automatic speech recognition (ASR) with the direct mapping of acoustic feature sequence to (sub)word sequence has shown great simplicity and state-of-the-art performance \cite{zoltan2020swb, Gulati20conformer}. Common end-to-end approaches include connectionist temporal classification (CTC) \cite{graves2016ctc}, recurrent neural network transducer (RNN-T) \cite{graves2012sequence}, attention-based encoder-decoder models \cite{bahdanau2016end, chan2016listen} and possible variants thereof. 

Similar as in natural language processing (NLP), subword units are commonly used for end-to-end ASR tasks to alleviate the out-of-vocabulary (OOV) issue as opposed to word-based systems.
In contrast to single character-based systems, subwords can avoid too long output sequence and dependency, which reduces the difficulty of modeling and decoding.
One popular subword modeling approach is the byte pair encoding (BPE) originally proposed for neural machine translation (NMT) \cite{sennrich16BPE}. Given a text corpus with all words split into single characters, pairs of units are merged based on frequency until a predefined vocabulary size is reached. 
A similar approach is the WordPieceModel (WPM) \cite{schuster12wordPiece}, where the subword merging is based on the likelihood of the text data. 
While both BPE and WPM lead to deterministic segmentation of words, \cite{kudo18uniLM} proposed the unigram language model (ULM) approach for subwords generation as well as a probabilistic segmentation. Starting with some seed vocabulary of subwords, the ULM approach trains the model via the EM algorithm w.r.t. a marginal likelihood over all possible within-vocabulary segmentation of the text data. Then vocabulary refinement and model training are iteratively repeated until a predefined vocabulary size. 
Based on the trained ULM, subword regularization is applied by drawing samples of segmentation variants in training NMT models, which is shown to improve performance \cite{kudo18uniLM}. 
However, all of these approaches are purely text-based without consideration of the underlying acoustic signal which is the key of ASR.

While automatic learning of label units from an acoustic perspective has been well studied for classical ASR systems \cite{michelPhDthesis}, it is not fully addressed in end-to-end ASR.
The pronunciation-assisted subword modeling (PASM) approach \cite{xu19pasm} tries to extract subword units by exploring their acoustic structure based on a pronunciation lexicon. The final labels are obtained with some post-processing based on a text corpus. However, no acoustic data is involved in the process and the label quality can be largely influenced by the underlying grapheme-to-phoneme (G2P) aligning toolkit.
Both the latent sequence decompositions (LSD) approach \cite{chan17lsd} and the GramCTC approach \cite{liu17gramCTC} utilize the acoustic data for subword decomposition.
The general idea is to improve ASR performance with a modified loss function that exposes the ASR model to various segmentations in training, so that the model jointly learns an acoustic-based sequence decomposition within a fixed subword vocabulary.
This is realized by applying sampling with the model in training for the LSD approach, and by marginalization over all segmentation variants for the GramCTC approach.
However, neither of them aims for a fully acoustic-oriented subword modeling, which is already clear from the simple vocabulary initialization with the most frequent $n$-gram characters in the transcription. Although both approaches are able to generate some subword set, the effect on further end-to-end ASR using different models is also missing.

In this work, we propose an acoustic data-driven subword modeling (ADSM) approach that combines most advantages of the aforementioned methods. 
Similar as PASM, the ADSM labels are designed with acoustic structure from the very beginning.
We use the GramCTC loss to train models on the acoustic data for vocabulary refinement in an acoustic data-driven manner. 
Additionally, subword merging as in BPE or WPM is applied to further enhance the vocabulary with larger units while retaining the acoustic correspondence. And the selection of smaller or larger units is decided in the next iteration of model training to again refine the vocabulary based on the acoustic input. The final iteration reveals the acoustic-oriented subword labels together with the acoustic-matched target sequence for further ASR training. 
The obtained ADSM labels are further evaluated on different end-to-end ASR approaches including CTC, RNN-T and attention models. Experiments on the LibriSpeech corpus \cite{libsp} show that ADSM clearly outperforms both BPE and PASM in all cases.

\section{Proposed Approach}
\subsection{Model}
Let $\vv{a}$ denote a sequence of subwords $a$ from a vocabulary $V$, and let $S(w)=\{\vv{a}:w\}$ define the set of allowed segmentation of the word $w$ using $a \in V$.
For each training utterance, let $X$ be the input acoustic feature sequence and $W$ be the corresponding word sequence. We denote $A \in S(W)$ as the allowed subword sequence for the full utterance, where no cross-word segmentation is considered. 
The training objective is then defined as:\\
\scalebox{0.9}{\parbox{1.11\linewidth}{%
\begin{align*}
\mathcal{L}(\theta) = -\log \sum_{A \in S(W)} p(A \mid X; \theta) \numberthis \label{eq:loss}
\end{align*}}}
where $\theta$ stands for the underlying model parameters. Note that this loss function can also be regarded as an extension of the marginal likelihood defined in the ULM approach \cite{kudo18uniLM} with further dependency on the acoustic input.

Let $h_1^T = f_{\theta}^{\text{enc}}(X)$ denote the model's encoding transformation which also includes optional subsampling. Similar as the GramCTC \cite{liu17gramCTC}, we apply an extended CTC approach to further define the loss in \Cref{eq:loss} as:\\
\scalebox{0.9}{\parbox{1.11\linewidth}{%
\begin{align*}
\mathcal{L}(\theta) &= -\log \sum_{A \in S(W)} \sum_{y_1^T:A} p'(y_1^T \mid h_1^T; \theta)\\
&= -\log \sum_{A \in S(W)} \sum_{y_1^T:A} \prod_{t=1}^T p'(y_t \mid h_1^T; \theta) \numberthis \label{eq:lossCTC}
\end{align*}}}
where $y_1^T$ is the blank $\epsilon$-augmented CTC alignment sequence of $A$ and $p'$ is defined over $V \cup \{\epsilon\}$. 
The unique mapping $B(y_1^T) = A$ is done via the CTC collapsing function $B$ which removes all label loop and blank $\epsilon$ in $y_1^T$. This loss function effectively computes a marginalization over all CTC alignments of all allowed subword decomposition of $W$.

Based on \Cref{eq:lossCTC}, we can train models on the acoustic training data as used for ASR training. 
This allows the model to learn more probable segmentations of each utterance in an acoustic data-driven manner.
The resulting probability distribution can capture the variation of the acoustic input and may favor different subword sequences for the same word depending on the audio. Additionally, a certain degree of the transcription statistics is also implicitly accounted in training, where those subwords belonging to the more frequent words are simply seen more often.

\subsection{Initialization}
Our vocabulary initialization is based on the lexicon, which is easily accessible and contains carefully designed expert knowledge on pronunciations of each word. By using a common G2P mapping toolkit such as the Phonetisaurus \cite{novak16phonetisarus}, we obtain aligned pairs of subword and phoneme for each word in the lexicon. 
If a subword is aligned to the null symbol, we merge it to the neighboring subword that has phoneme mapping. 

Unlike PASM \cite{xu19pasm}, we do not directly rely on these G2P pairs as they may contain errors. Instead, we simply take all the subword units from these pairs and use them as our initial vocabulary $V$. Most of these subword units can be regarded as graphemic representation of phonemes and thus, have a clear correspondence to certain acoustic structure.
Then $S(w)$ is defined as all possible segmentation of the word $w$ using $a \in V$, which also relaxes the quality requirement of the G2P alignment.
Additionally, subwords appearing at word end are regarded as different labels, i.e. $a\_$, since they might have different acoustic property \cite{le2019chenones}.
This word-end distinction also allows a straightforward reconstruction of words and is further carried over into $V$ and $S(w)$.
Some examples of $S(w)$ after this step are shown in the second column of \Cref{tab:example}.
We use an initial subsampling of factor 2.

\subsection{Iteration}
\subsubsection{Step $1$: vocabulary refinement}
Given $V$ and $S(w)$, a new CTC model can be trained based on \Cref{eq:lossCTC}, where the model's NN architecture is free to choose. One can simply use the same encoder network as further ASR models to have a more consistent behavior. 
With this trained model, we apply vocabulary refinement by simply performing a Viterbi aligning on the acoustic training data. For each utterance $(X, W)$, the optimal subword sequence can be obtained as:\\
\scalebox{0.9}{\parbox{1.11\linewidth}{%
\begin{align*}
\tilde{A} &= B(\hspace{1.5mm} \argmax_{y_1^T:A\in S(W)} \frac{p'(y_1^T \mid h_1^T; \theta)}{q^{\lambda}(y_1^T)} \hspace{1.5mm})\\
&= B(\hspace{1.5mm} \argmax_{y_1^T:A\in S(W)} \prod_{t=1}^{T} \frac{p'(y_t \mid h_1^T; \theta)}{q^{\lambda}(y_t)} \hspace{1.5mm})
\end{align*}}}
where $q$ is the prior distribution estimated by marginalizing $p'$ over the complete training data. By adjusting the scale $\lambda \in [0, 1]$, we can control the smoothness of the overall model for this aligning process. An increasing $\lambda$ leads to more segmentation variants of each word in the alignment.

Based on the forced alignment of the complete training data, we gather all subword sequence variants of each word, i.e. $\vv{a}$, together with their counts. These counts are then renormalized to weights by the total occurrence of that word. We introduce a threshold $\mu$ to perform a per-word $\vv{a}$ filtering. Namely, for each word, we filter out its subword sequence variants whose weight is less than $\mu$. Finally, the remaining $\vv{a}$ of each word $w$ form a new $\tilde{S}(w)$ as well as a new vocabulary $\tilde{V}$. Here $\lambda$ and $\mu$ jointly control the size of $\tilde{V}$. 
Some examples of $\tilde{S}(w)$ at this step after initialization are shown in the third column of \Cref{tab:example}. 
Note that special care should also be taken to ensure that all single characters are included in the vocabulary. However, we empirically find that this is always achieved with the above procedure, which is probably due to the rich statistics of single characters.

\begin{table}[t!]
\caption{\it Example of subword segmentation for the words ``able" and ``word" after specific ADSM steps}
\centering\label{tab:example}
\setlength{\tabcolsep}{0.5em}
\begin{tabular}{|c|c|c|c|c|}
\hline
\multirow{2}{*}{Word} & \multirow{2}{*}{Initialization} & Vocab- & Subword- & Vocab- \\
                      &      & refinement & merging & refinement \\ \hline
\multirow{3}{*}{able} & a b l e\_ & \multirow{3}{*}{a ble\_} & \multirow{3}{*}{\shortstack[c]{a ble\_\\able\_}} & \multirow{3}{*}{a ble\_} \\
                      & a b le\_  & & & \\
                      & a ble\_   & & & \\ \hline
\multirow{4}{*}{word} & w o rd\_ & \multirow{4}{*}{w or d\_} &\multirow{4}{*}{\shortstack[c]{w or d\_\\w ord\_\\wor d\_}}& \multirow{4}{*}{\shortstack[c]{w or d\_\\w ord\_}} \\
                      & w or d\_ &&& \\
                      & wo r d\_ &&& \\
                      & wo rd\_ &&& \\
\hline
\end{tabular}
\vspace{-3mm}
\end{table}

\subsubsection{Step $2$: subword merging}
One major idea of several text-based subword approaches is the merging of subword pairs based on certain criterion, such as frequency for BPE \cite{sennrich16BPE} and data likelihood for WPM \cite{schuster12wordPiece}. This can ease the burden of learning spelling with many small units and benefit context-dependency modeling. Following this idea, we also apply subword merging adapted into this framework. For each $\vv{a}$ in $\tilde{S}(w)$ of each word, we merge any two neighboring units to form all possible new sequences. For example, merging the sequence $\vv{a}=(a_1,a_2,a_3,a_4)$ leads to $(a_1a_2, a_3, a_4), (a_1,a_2a_3, a_4)$ and $(a_1,a_2, a_3a_4)$. The merged unit becomes a new label in $\tilde{V}$ and the new sequence updates $\tilde{S}(w)$ as well. Note that the original sequence before merging is always kept in $\tilde{S}(w)$. Since all subword units before merging correspond to certain acoustic structure, the merged unit also retains this correspondence. 

With the updated $\tilde{V}$ and $\tilde{S}(w)$, we can repeat the iteration to train a new model that again learns proper decomposition using the new labels in an acoustic data-driven manner. Additionally, we further increase the subsampling factor by 2 to encourage the model to capture larger units.
Some examples of $\tilde{S}(w)$ after the subword merging and subsequent vocabulary refinement steps are shown in the last two columns of \Cref{tab:example}.
Theoretically, this procedure can be repeated many times. However, we find that it is sufficient to repeat this just once or twice, since further merging results in many very long and rare units that suffer the data sparsity problem.

\subsection{Finalization}
\subsubsection{Label and target sequence}
The finalization is done together with one last vocabulary refinement step. 
To further eliminate subword units of insufficient statistics, we introduce another word-count threshold $k$ for $S_{\text{final}}(w)$. Namely, if a word occurs less than $k$ times in the data, we only take its best $\vv{a}$ based on weights. With $S_{\text{final}}(w)$, the final subword vocabulary $V_{\text{final}}$ can be obtained, whose size is controlled by $\lambda$, $\mu$ and $k$ jointly. Starting from the initialization, the whole process of label design and learning is fully acoustic oriented to better serve the task of ASR. 
By replacing those filtered subword sequences with the best alternative in $S_{\text{final}}(w)$, the final forced alignment directly reveals the target sequence for further ASR training. This can be regarded as the acoustically most probable decomposition of each utterance.

\subsubsection{Subword regularization}
Additionally, we can also disregard the obtained target sequence and $S_{\text{final}}(w)$. Then based on $V_{\text{final}}$ only, different segmentation variants for the same word appear, where subword regularization can be applied. 
The ULM-based subword regularization \cite{kudo18uniLM} is shown to bring some improvement on ASR performance \cite{drexler19swRegE2Easr, hannun19swRegS2S, lakomkin20swRegAmazonASR}. However, it is not guaranteed that subword samples drawn from a text-based probabilistic model can always fit the underlying acoustic signal.
In ADSM, we can simply use the final model or train a new model on $V_{\text{final}}$ to perform acoustic-based subword regularization in ASR training. Thus, the subword samples obtained are more consistent with the ASR objective.

\subsubsection{Text segmentation}
Segmentation of text without corresponding audio is needed for training a subword LM on extra text data. This can be simply realized with ADSM based on statistics from an acoustic preference. 
For words in $S_{\text{final}}(w)$, we draw samples of $\vv{a}$ based on weights. For unseen words, we train an $n$-gram LM on the complete $S_{\text{final}}(w)$ to score the best segmentation among all possible variants within $V_{\text{final}}$.
Since all single characters are included in $V_{\text{final}}$, a fully open-vocabulary system is always guaranteed.

\section{Experiments}
\subsection{Setup}
Experimental verification is done on the 960h LibriSpeech \cite{libsp} corpus. By default, we use 50-dimensional gammatone features from 25ms windows with 10ms shift.
For the ADSM training, we use 6 bi-directional long short-term memory (BLSTM) \cite{hochreiter1997lstm} layers with 512 units for each direction and a softmax output. The increasing subsampling is done via adding max-pooling layers into the NN. We use the nadam optimizer \cite{nadam} with an initial learning rate (LR) 0.001 and a minimum LR $10^{-5}$. The layer-wise pretraining \cite{bengio06layerwisePretrain} and the newbob LR scheduling \cite{zeyer2017newbob} with a decay factor 0.9 are applied. 
Additionally, we also include SpecAugment \cite{zoph2019specaugment} with a gentle masking setting as in \cite{zhou2020icassp} to improve the model's generalization. 
An efficient Baum-Welch CUDA implementation as described in \cite{zeyer17ctc} is used.
Each iteration of model training converges well within about 25 full epochs, which can be done on a single GTX-1080-Ti-GPU in about one week.
The best epoch to perform the aligning step is selected based on the cross-validation score and some intermediate recognition.

The obtained ADSM labels are further evaluated against both BPE \cite{sennrich16BPE} and PASM \cite{xu19pasm} for subword-based end-to-end ASR. The comparison is done on different end-to-end models including CTC \cite{graves2016ctc}, monotonic RNN-T \cite{tripathi2019monoRNNT} and LSTM-based attention model \cite{zeyer2019attention}.
The CTC model uses the same NN architecture and similar training setup as ADSM. The RNN-T model uses the ADSM NN structure for the encoder, 2 LSTM layers with 1024 units for the prediction network and a simple addition for the joint network. Here we follow a Viterbi training variant for transducer models as described in \cite{zhou2021phonemeTransducer}, which is shown to be both better and more efficient than standard full-sum training under the same limited number of epochs. 
The attention model is closely following \cite{zeyer2019attention} in terms of the NN architecture, training pipeline and the use of MFCC features, which are originally optimized on the BPE units.
SpecAugment is applied in all training and a maximum of 30 full epochs is used. 

For fair comparison with BPE and PASM, we only use the ADSM target sequence for training without applying subword regularization.
Although possible, we also explicitly do not use the ADSM model to initialize the end-to-end models.
For each model type, all three subword units are trained under the same setup with minimum tuning effort. 
For recognition, we use a beam size in $\{64, 128\}$ and no additional LM is used.

\begin{table}[t!]
\caption{\it Subword vocabulary size, average number of subword sequence variants per word, and average length of subword sequences of all words after each ADSM step}
\centering\label{tab:ADSM}
\setlength{\tabcolsep}{0.5em}
\begin{tabular}{|c|c|r|r|c|}
\hline
\multicolumn{2}{|c|}{\multirow{2}{*}{Step}} & \multirow{2}{*}{ $|V|$} & \multicolumn{2}{|c|}{Average} \\ \cline{4-5}
  \multicolumn{2}{|c|}{}                    & & { $|S(w)|$} & { len$(\vec{a})$ } \\ \hline
\multicolumn{2}{|c|}{Initialization} & 2k & 51.7 & 8.1\\ \hline
\multirow{2}{*}{1 Iteration} & vocab-refinement & 1k & 1.2 & 5.4 \\ \cline{2-5}
& subword merging & 21k & 6.4 & 5.2 \\ \hline
\multicolumn{2}{|c|}{Finalization} & \textbf{5k} & \textbf{1.1} & \textbf{4.7} \\ \hline
\end{tabular}
\vspace{-3mm}
\end{table}

\subsection{ADSM}
The initialization step is done based on the official lexicon of LibriSpeech. The complete iteration of training and aligning for vocabulary refinement as well as subsequent subword merging is applied only once. Then we directly move to the final iteration of vocabulary refinement with finalization, which reveals the final ADSM labels and target sequence. Throughout this procedure, we use $\lambda=0.3$, $\mu=0.05$ and $k=20$.

\Cref{tab:ADSM} shows the concrete vocabulary size $|V|$, the average number of subword sequence variants per word $|S(w)|$ and the average length of all subword sequences len$(\vv{a})$ after each ADSM step. A clear reduction of $|V|$ and $|S(w)|$ can be seen before and after each vocabulary refinement step, which indicates the model's convergence towards specific acoustic-probable decomposition of each utterance.
The small value of converged $|S(w)|$ approaches the ratio of pronunciation variants in the original LibriSpeech lexicon, which suggests only a small fraction of segmentation variants in the target sequence. This is intuitively clear since the ADSM units are designed from phoneme correspondence and have a more deterministic relation with the audio.
A final ADSM vocabulary size of 5k is obtained, which is reasonable for the LibriSpeech corpus. For fair comparison, we also generate BPE and PASM subword sets of the same size. 
Additionally, the model also learns to capture larger units as shown by the decreasing len$(\vv{a})$. For reference, the average pronunciation length with phonemes is 6.5, while BPE and PASM have a corresponding length of 3.2 and 5.7, respectively.

\begin{table}[t!]
\caption{\it WER results on the LibriSpeech datasets for different subword units under different end-to-end ASR approaches (without external LM)}
\centering\label{tab:WER}
\setlength{\tabcolsep}{0.5em}
\begin{tabular}{|c|c|c|c|c|c|}
\hline
\multirow{2}{*}{Model} & \multirow{2}{*}{Subword} & \multicolumn{2}{c|}{dev} & \multicolumn{2}{c|}{test} \\ \cline{3-6}
& & clean & other & clean & other \\ \hline \hline
\multirow{3}{*}{CTC} & PASM & 9.0 & 21.2 & 8.9 & 21.5 \\ \cline{2-6}
                     & BPE & 9.5 & 20.0 & 9.5 & 20.9 \\ \cline{2-6}
                     & ADSM & \bf{8.7} & \bf{20.0} & \bf{8.7} & \bf{20.6} \\ \hline \hline
\multirow{3}{*}{RNN-T} & PASM & 5.3 & 13.2 & 5.4 & 13.6 \\ \cline{2-6}
                     & BPE & 5.6 & 13.2 & 5.9 & 14.0 \\ \cline{2-6}
                     & ADSM & \bf{5.0} & \bf{12.6} & \bf{5.2} & \bf{12.8} \\ \hline \hline
\multirow{3}{*}{Attention} & PASM & 4.9 & 13.5 & 5.2 & 14.5 \\ \cline{2-6}
                           & BPE & 4.9 & 13.0 & 5.1 & 13.6 \\ \cline{2-6}
                           & ADSM & \bf{4.8} & \bf{12.8} & \bf{5.0} & \bf{13.5} \\ \hline
\end{tabular}
\vspace{-3mm}
\end{table}

\subsection{End-to-end ASR}
Together with BPE and PASM, we evaluate the effect of the obtained ADSM labels on three popular end-to-end ASR approaches, i.e. CTC, RNN-T and attention models. The word error rate (WER) results are shown in \Cref{tab:WER}, where ADSM outperforms both BPE and PASM in all cases.
For the CTC model, ADSM brings large improvement on the clean-sets and small improvement on the other-sets. 
Without any context dependency, the acoustic correspondence of subword units and the sequence length, e.g. difficulty of spelling, become critical. The latter is directly related to the label size, where PASM seems to suffer the much longer sequence length.
The biggest improvement of ADSM is observed for the RNN-T model. While the sequence length issue is largely addressed by the context dependency, both ADSM and PASM outperform BPE units, which indicates the benefit of labels' acoustic structure for such time-synchronous model.
The difference between ADSM and BPE becomes smaller for the attention model, although we observe an easier convergence for ADSM at the beginning of training. Besides the inconsistent setup with ADSM training, we infer that the clear acoustic correspondence of subword units is less critical for such label-synchronous model. 
In this case, the context dependency has an increasing importance for the attention mechanism, where text-based BPE has a clear advantage and PASM again has the disadvantage of much longer sequence length. Overall, ADSM achieves the best performance for both time-synchronous and label-synchronous models.

\subsection{Analysis}
To have more insights of the performance difference, we also check into more details of these subword approaches in terms of acoustic meaningfulness and label size. The latter in turn affects the overall sequence length. \Cref{tab:example} shows some typical examples.
As a purely text-based approach, BPE segmentation can sometimes completely disregard the acoustic structure of words' pronunciation, which might increase the difficulty of ASR modeling. 
While PASM alleviates this issue, the resulting subword decomposition is not always optimal. 
As an acoustic data-driven approach, ADSM produces more acoustically logical subword units as well as decomposition of words, which can be easier for the ASR model to capture from the acoustic input.

The label size is mostly decided by the approach design and larger labels lead to shorter output sequences. 
The BPE units are usually large since frequent subword pairs are simply merged. 
As a result, many frequent words become single labels and less frequent words are decomposed into much smaller units, which certainly has an advantage on context dependency modeling.
Without any merging operation, the PASM labels are usually small, although some large units can also occur biased by the underlying G2P aligning toolkit. 
The ADSM labels have an acoustically more balanced length, which leads to a more balanced overall sequence length.
All these can also have a joint effect with the external subword LM, which we will investigate in future work.

To illustrate the importance of both acoustic structure and label size, we perform an idealized recognition using the subword CTC models and the official LibriSpeech 4-gram word-level LM. 
Here the spelling is perfectly defined in the dictionary and cross-word context is captured by the LM. 
As shown in \Cref{tab:ctcLM}, both acoustic-based subword units become similarly good and outperform BPE. 
In contrast to the CTC results in \Cref{tab:WER}, PASM has the most significant gain in this ideal case, which partially justifies the disadvantage of too long sequence length for end-to-end ASR.

\begin{table}[t!]
\caption{\it Example of segmentation using different subwords}
\centering\label{tab:example}
\setlength{\tabcolsep}{0.5em}
\begin{tabular}{|c|l|l|l|}
\hline
Subword  & ``bachelor''  & ``password''  & ``together'' \\ \hline
PASM     & b a ch elor\_ & p a s s w or d\_ & togethe r\_ \\ \hline
BPE      & bac hel or\_  & pas sword\_  & together\_ \\ \hline
ADSM & b a chel or\_ & p a ss w ord\_ & to g e ther\_ \\ \hline
\end{tabular}
\vspace{1mm}
\end{table}

\begin{table}[t!]
\caption{\it WER results of the CTC models with an external 4-gram word-level LM on the  LibriSpeech datasets}
\centering\label{tab:ctcLM}
\setlength{\tabcolsep}{0.5em}
\begin{tabular}{|c|c|c|c|c|}
\hline
\multirow{2}{*}{Subword} & \multicolumn{2}{c|}{dev} & \multicolumn{2}{c|}{test} \\ \cline{2-5}
 & clean & other & clean & other \\ \hline
 PASM & 4.1 & 10.4 & 4.3 & 10.9 \\ \hline
 BPE  & 4.7 & 11.2 & 4.8 & 11.9 \\ \hline
ADSM  & 4.1 & 10.2 & 4.6 & 11.0 \\ \hline 
\end{tabular}
\vspace{-3.5mm}
\end{table}

\section{Conclusions}
In this work, we presented an acoustic data-driven subword modeling (ADSM) approach that adapts the advantages of several text-based and acoustic-based subword methods into one detailed pipeline. 
The complete procedure of label design and learning is fully acoustic-oriented and thus, more consistent with the ASR objective.
ADSM produces the acoustic-structured subword units together with the acoustic-matched target sequence for further ASR training. The target sequence can be regarded as the acoustically most probable decomposition of each utterance.
The obtained ADSM labels are further evaluated on different end-to-end ASR approaches including CTC, RNN-T and attention models. Experiments on the LibriSpeech corpus show that ADSM clearly outperforms both BPE and PASM in all cases. 
Detailed analysis shows that ADSM achieves acoustically more logical word segmentation and more balanced sequence length, and thus, is suitable for both time-synchronous and label-synchronous models.
We also briefly described how to apply acoustic-based subword regularization and unseen text segmentation using ADSM.

Future work includes the investigation of the joint effect with an external subword LM, and the inclusion of more text statistics into the label learning process, which aims towards a joint acoustic and textual data-driven subword modeling.

\section{Acknowledgements}
\scriptsize
This work has received funding from the European Research Council (ERC) under the European Union's Horizon 2020 research and innovation programme (grant agreement No 694537, project ``SEQCLAS'') and partly from a Google Faculty Research Award (``Label Context Modeling in Automatic Speech Recognition''). The work reflects only the authors' views and none of the funding parties is responsible for any use that may be made of the information it contains.

\bibliographystyle{IEEEtran}
\bibliography{refs}

\end{document}